\pdfoutput=1

\documentclass[11pt]{article}

\usepackage{acl}
\usepackage{times}
\usepackage{latexsym}

\usepackage{graphicx}

\usepackage{subfigure}
\usepackage[T1]{fontenc}

\usepackage[utf8]{inputenc}

\usepackage{microtype}
\usepackage{makecell}
\title{Using Large Language Model to Solve and Explain Physics Word Problems Approaching Human Level}

\author{Jingzhe Ding \\
  Columbia University\\
  \texttt{jd4001@columbia.edu} \\\And
  Yan Cen \ Xinyuan Wei \\
  Fudan University\\
  \texttt{\{yancen,xinyuanwei\}@fudan.edu.cn}
  }

\begin{document}
\maketitle
\begin{abstract}
Our work demonstrates that large language model (LLM) pre-trained on texts can not only solve pure math word problems, but also physics word problems, whose solution requires calculation and inference based on prior physical knowledge. We collect and annotate the first physics word problem dataset \textbf{PhysQA}, which contains over 1000 junior high school physics word problems (covering Kinematics, Mass\&Density, Mechanics, Heat, Electricity). We then use OpenAI's GPT3.5 to generate answers to these problems and found that GPT3.5 could automatically solve 49.3\% of the problems through zero-shot learning and 73.2\% through few-shot learning. This result demonstrates that by using similar problems and their answers as prompts, LLM can solve elementary physics word problems approaching human-level performance. In addition to solving problems, GPT3.5 can also summarize the knowledge or topics covered by the problems, provide relevant explanations, and generate new physics word problems based on the input. Our work is the first research to focus on the automatic solving, explanation, and generation of physics word problems across various types and scenarios, and we achieve an acceptable and state-of-the-art accuracy. This underscores the potential of LLMs for further applications in secondary education.
\end{abstract}

\section{Introduction}

As one of the most focused areas in natural language processing, automatic solution of Math word problem (MWP) reflects the capability of artificial intelligence \cite{roy2015solving,zhang2019gap,patel2021nlp}. Until our work,almost all the research of automatic solution of math word problem using deep learning methods concentrate on pure math problems. With the development of the GPU and large language models (LLMs) \cite{zhao2023survey}, MWP solver could work out more difficult types of problems including geometry problems, advanced mathematics problems,etc.

\begin{figure}
\centering
\includegraphics[height=6cm]{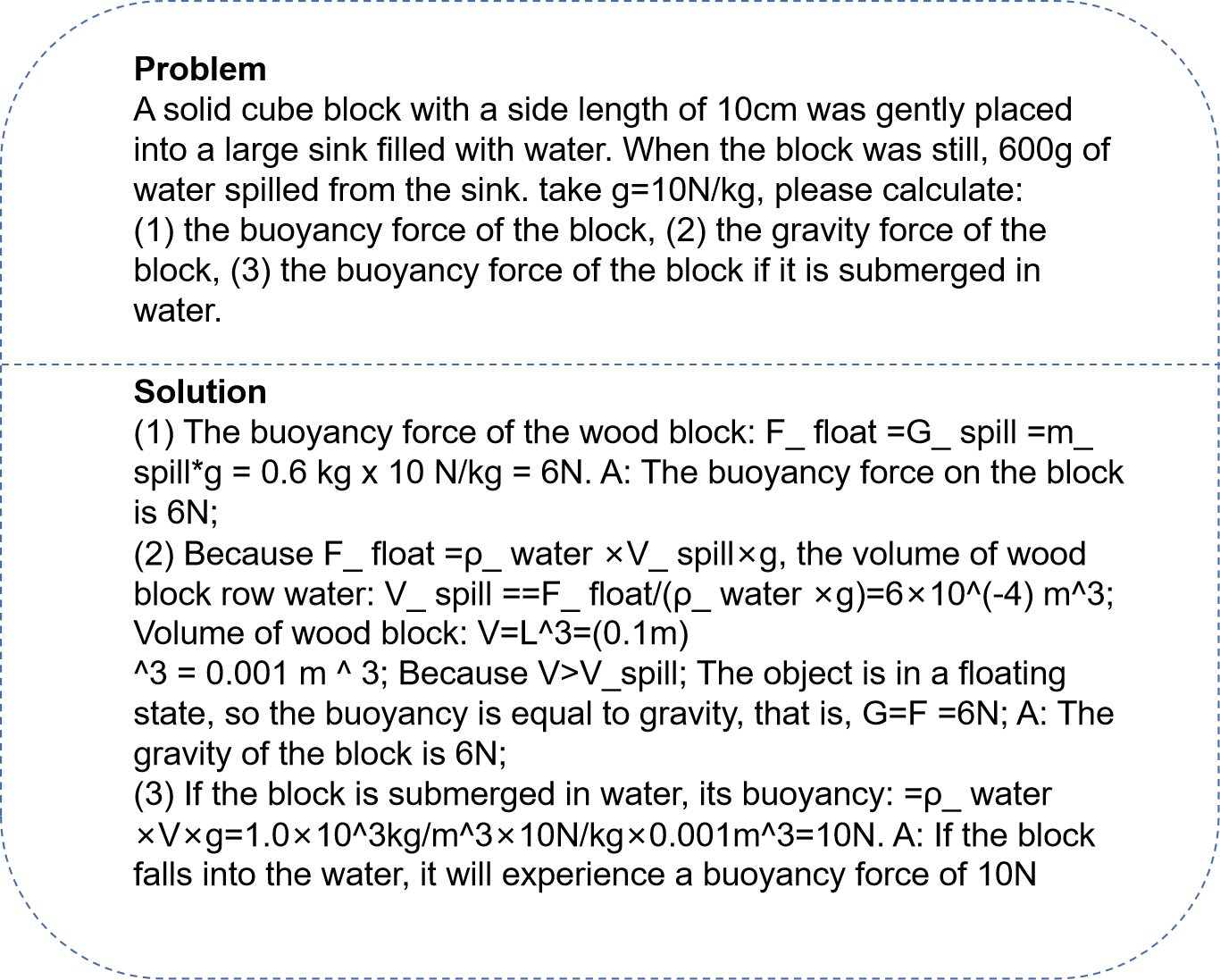}
\caption{Illustration of a typical example of physics word problem containing multiple sub-questions. One have to apply relevant physical formula and make reasoning to solve it.}
\label{fig1}
\end{figure}

However, physics word problems, which refer to questions to be answered by applying not only calculation,but also prior physical knowledge, also hold significant importance in all levels of education. Physics word problems primarily assess students' understanding of physical knowledge, analytical abilities, and mathematical skills. Students' ability to solve these problems can impact their characteristics and overall development \cite{abraham2016hybrid,ince2018overview}. As shown in Figure \ref{fig1}, a typical physics word problem typically describes a specific physical scenario and requires the calculation of certain variables based on provided information.

Compared with pure math word problem, physics word problem exhibits the following characteristics:

\noindent\textbf{The application of prior physical knowledge and formula} \  The fundamental prerequisite for accurately solving physics word problems is to have a precise understanding of physical properties and their dimensions, and then apply them correctly in the relevant formulas. For instance, when calculating mass\&density problems,one may utilize the formula \( \rho= \frac{m}{V} \).Similarly, in circuit problems, the widely used formula is Ohm's Law:\( I= \frac{U}{R} \).

\noindent\textbf{Containing more reasoning} \  For some physics questions, one have to make some inference from the text to accurately analyse the scene before performing calculations. As illustrated in Figure \ref{fig1}, it is necessary to infer from sub-question (1) that the volume of the fluid displaced by the wooden block is less than the block's own volume, so that the wood block is floating and the value of its gravity is equal to its buoyancy.

\noindent\textbf{More than one sub-questions} \  Most physics word problems depict specific physical scenarios and require individuals to calculate relevant physical variables. As shown in the problem in Figure \ref{fig1}, many scenarios involve multiple variables to be calculated, often leading to several sub-questions.

Due to these characteristics, physics word problems are actually more difficult and challenging than pure math word problems. However, despite the significance of physics in elementary education, there have been few efforts  focusing on the automatic solution of physics word problems, mainly due to the lack of relevant datasets.

To bridge this gap, we initially gathered 1008 distinct junior-high-school level problems from the internet, encompassing a total of 1770 sub-questions. Subsequently, we standardized their format, ensuring that each problem included essential information such as ID, solution, number of sub-questions, formulas, equations, answers, types, most similar problems, etc.This dataset marks the first collection of physics word problems, which we have named \textbf{PhysQA}. It serves as a valuable resource for future research in the field of physics word problems.

After completing our dataset, we apply the large language model GPT3.5 and ask it to solve the problems in PhysQA under different conditions--zero-shot learning and few-shot learning. In the few-shot learning part, we used two different types of prompt: paragraphs of textbook and similar problems.During these processes we find that similar questions(with solutions) greatly improve the model's performance, while paragraphs of textbooks have little influence on the total accuracy of all the problems. Moreover, we carry out some auxiliary tasks on the model like generating new problems, summarizing the core knowledge point or formula examined by the problems,etc. The result shows that although the model could finish these tasks, there are still some challenges and limitations.

In summarize,our main contributions are:
\begin{itemize} 
\item Collect and standardizdly annotate the first junior-high-school-level physics word problem dataset--\textbf{PhysQA},which contains 1008 questions(1770 sub-questions) of different types, and each problem includes comprehensive information.

\item Use GPT3.5 to solve the problems in PhysQA by zero-shot learning and few-shot learning and achieve the total accuracy of 73.2\% in few-shot learning. These results, along with those of other auxiliary tasks, demonstrate that large language models (LLMs) possess the capability to solve, explain, and generate physics word problems at a level approaching human performance.
\end{itemize}

\section{Related Work}

\noindent\textbf{Research on Automatic Solution of Physics Word Problem} \ Previous research on physics word problems has primarily focused on solving problems within fixed physical models or scenarios that involve minimal inference, such as free-falling objects under constant gravitational acceleration or the motion of objects under constant acceleration \cite{leszczynski2016machine,bleiweiss2019neural}. The limitations of these works are that they apply to restricted scenes and specific formulas,and mainly rely on predetermined problem formation and expression patterns. However, real-world physics word problems are not only flexible in their descriptions but also diverse in terms of models and scenarios. For a practical physics problem solver, the model should predict the knowledge points, formulas, and inferences contained in problems, rather than having them predefined.

\noindent\textbf{Using LLM to solve math word problem} \  Prior to the widespread adoption of large language models, most math word problem solvers were tree-structured or operation-based  \cite{wang2018translating,lu2019deeponet,2019A,koncel2015parsing,wang2018mathdqn,yang2022logicsolver,amini2019mathqa,chen2021geoqa}. In recent years, when the power of large language model (LLM) gradually appears, there has been a growing focus on applying LLMs to solve math word problems. The potential of LLM in few-shot learning also makes it possible to perform better in different tasks by fine-tuning or providing prompt \cite{brown2020language}. \citet{imani2023mathprompter} designed math-prompter method for LLM and improves over state-of-the-art on the MultiArith dataset from 78.7\% to 92.5\%.\citet{drori2022neural} used OpenAI’s Codex to solve University Math Problems at human level, and improves the benchmark accuracy form 18.8\% to 81.1\%. Additionally, the Xueersi team is actively working on developing a math-oriented model known as MathGPT.

Compared to traditional Math Word Problem (MWP) solvers, Large Language Models offer several advantages. The first advantage is interpretability. While non-LLM models have attempted to enhance interpretability by incorporating auxiliary tasks such as predicting formulas and relevant knowledge points, LLMs typically provide a clear train of thought, including the inference and calculation process, when solving problems. This makes it easier for users to understand the solutions. The second advantage is flexibility. To enhance the model's performance on specific problem types, we can provide useful prompts, a technique known as "Few-shot Learning" \cite{brown2020language}.

\noindent\textbf{The range of problems that could be solved automatically} \  Besides the well-known datasets like Math23K and Dolphin18K, which primarily consist of elementary-level math word problems \cite{wang2017deep,huang2016well}, some research works on the automatic solution of more challenging problems. \citet{chen2021geoqa} collected and annotated GeoQA---a large-scale dataset of geometric word problems and introduce a Neural Geometric Solver(NGS) to solve them. \citet{drori2022neural} focus on the automatic solution of advanced math word problems. 

However, as mentioned earlier, only few works focus on any difficulty level of physics word problems. 

\section{Dataset}

\subsection{Collect and Annotate PhysQA Dataset}
Compared with MWP, which has abundant datasets, there is no available physical problem dataset containing sufficient problems in similar difficulty level and covering multiple physical scenes. Although some LLM testing datasets like MMLU 
\cite{hendrycks2020measuring} contain some physics questions, these questions’ difficulties vary greatly, and sometimes they directly examine some certain formula or knowledge without the process of calculation or reasoning. So we construct our dataset by selecting physics word problem form some Chinese test paper websites like Canpoint\footnote{\url{https://ti.canpoint.cn/paper/index}} and Jyeoo\footnote{\url{http://www.jyeoo.com/}}. 

\begin{figure}
\centering
\includegraphics[height=5cm]{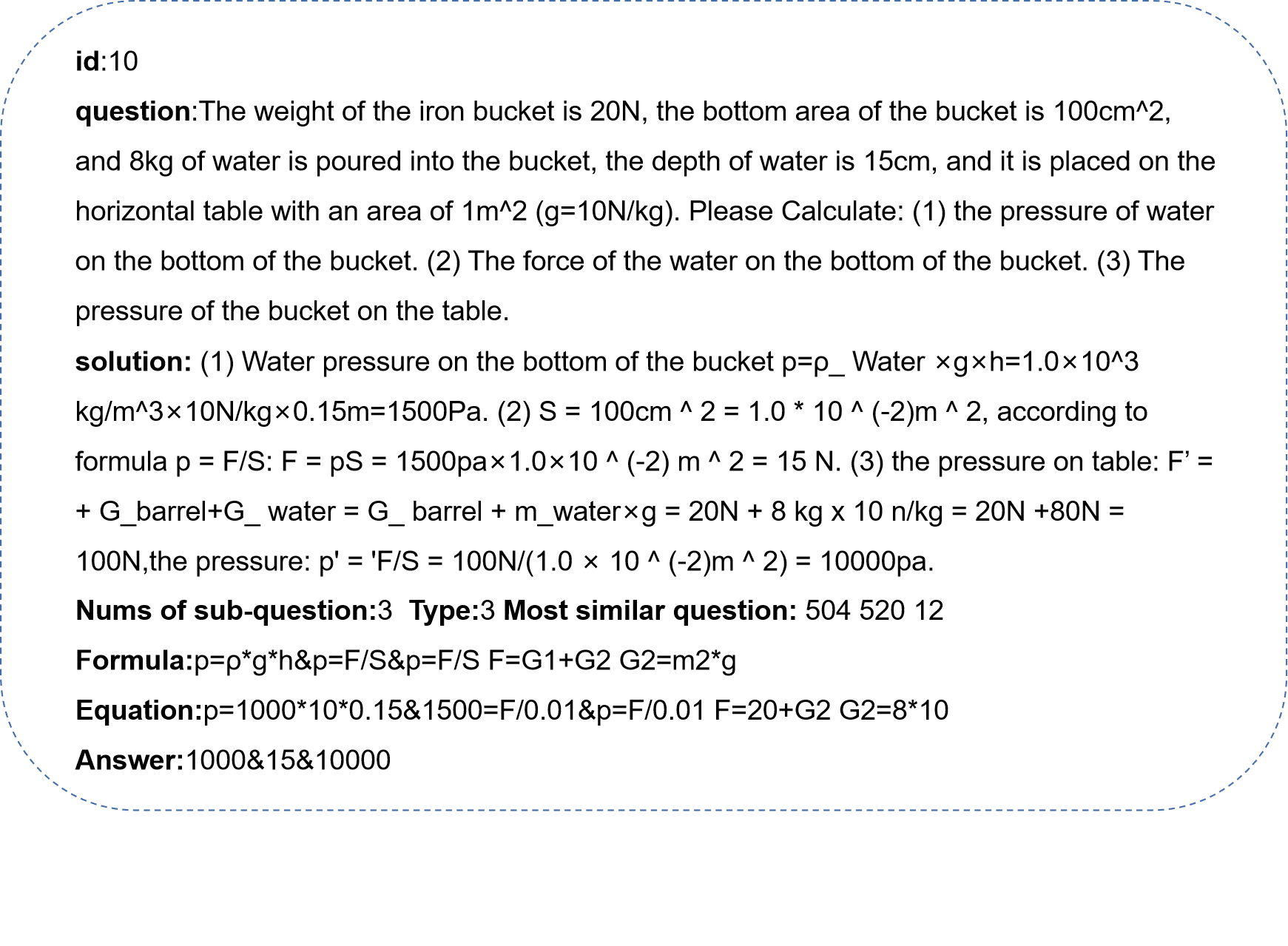}
\caption{An example of annotated problems in PhysQA dataset.}
\label{fig2}
\end{figure}

After downloading problems as well as their solutions from the website, we annotate them to a standard format. An example of annotated problem can be found at Figure \ref{fig2}. Considering GPT3.5 could not read figures, we only choose problems without pictures or the picture without extra information for the problem.

Finally we have completed a dataset containing 1008 junior high school physics word problems, categorized into 5 different types, with a total of 1770 sub-questions. Each problem has its id, text, solution, nums of question, most similar question, formula,equation and (numeric)answer. The detailed composition of dataset is shown in Table \ref{tab1}. This marks the first standardized and annotated dataset of physics word problems, which we have named \textbf{“PhysQA”}. Following our work, this dataset can be utilized for research related to the automatic solution of physics word problems. Since we annotated the formula, equation and (numerical) answer of the problems, our dataset is compatible with both large language models and tree-structured or operation-based physics problem solvers. 

Moreover, unlike the annotation of some previous datasets needing to hire volunteers,we annotate the data on our own with the assistance of OpenAI's large language model (gpt-3.5-turbo and text-davinci-003), especially in the part of "formula" and "equation".We find that although the model could pick out the formulas according to the solutions correctly, in most cases the form of its output couldn't meet our inputted demands. Even so it significantly reduces time spent on annotating dataset.

\begin{table*}
\centering
\resizebox{\textwidth}{!}
{
\begin{tabular}{ccccccc}
\hline
\textbf{Number} & \textbf{Kinematics} & \textbf{Mass\&Density}& \textbf{Mechanics} & \textbf{Heat}&\textbf{Electricity} & \textbf{Total}\\
\hline
problems & 299 & 175 &192 &277 &65 &1008 \\
sub-questions & 493 & 330 &344 &483 &120 &1770 \\
\hline
average sub-questions & 1.65 & 1.89 &1.79 &1.74 &1.85 &1.76 \\
\hline
\end{tabular}
}
\caption{
Composition of PhysQA dataset. It contains 1008 problems (1770 sub-questions) from 5 main areas of junior high school physics.
}
\label{tab1}
\end{table*}

\subsection{Pre-process Dataset}
When calculating the “most similar question”, we apply Sentence-BERT \cite{kenton2019bert,reimers2019sentence} pre-trained model to obtain the embeddings of each problem.To minimize the cost of computility , we specifically calculate cosin similarity of questions within each type. For each question, we select the top-3 most similar problems and record their IDs to our dataset in descending order of text similarity.

Addictionally, we found that for a very small proportion of our dataset, although some problems are identified as different on the website, their text of problems are actually all the same. As for these problems we make slight changes on them like replacing the number appeared or paraphrasing the problems.

\section{Experiments}
After completing our datastet, we apply OpenAI’s GPT3.5 (gpt-3.5-turbo) to solve these problems. We apply two methods:

\noindent\textbf{Zero-shot Learning:} \  Directly instruct model to solve the problem.

\noindent\textbf{Few-shot Learning:} \  In the case of few-shot learning, we primarily experiment with two methods to provide prompts to the model.

\subsection{Methods of Few-shot Learning}

\begin{figure}[htb]
\centering
\includegraphics[width=0.45\textwidth]{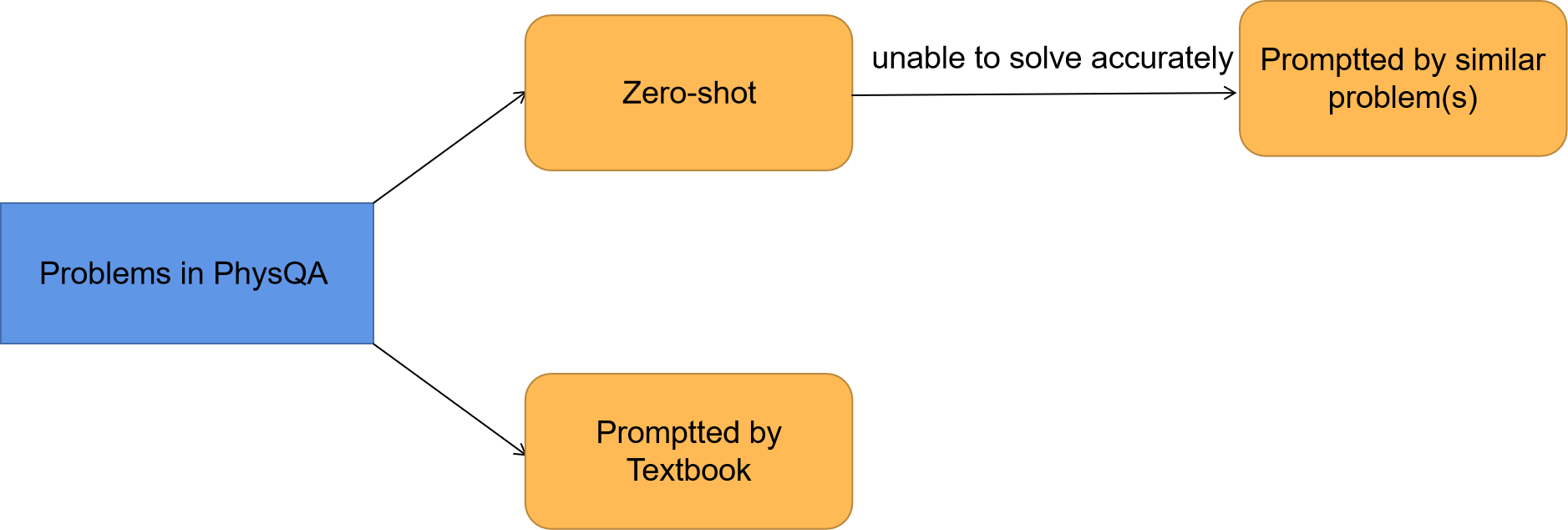}
\caption{All problems are initially inputted under the condition of zero-shot learning and prompted by textbooks. If the solution generated by zero-shot learning is inaccurate, similar problems and their solutions are used as prompts for resolution.}
\label{fig3}
\end{figure}

\textbf{Prompted by Similar Question} \  As shown in Figure \ref{fig3}, if the zero-shot answer generated by GPT3.5 is incorrect (unable to solve all sub-questions), we input the similar problems and solutions as prompt, and use the model to solve it again. Since we have calculated the indices of top 3 most similar questions, we experiment with using one and three prompt questions during the process.

\noindent\textbf{Prompted by Textbook} \ We extract some essential paragraphs(including the definition\& explanation of some physics quantity, the meaning of significant physics formula) from junior high school physics textbook. For each problem in PhysQA, we input the relevant paragraphs together with the problem and order the model to solve it. We apply this method to all the problems, irrespective of whether the zero-shot result is accurate.

\subsection{Test Problems Outside Dataset}
Over 90\% of problems in PhysQA are selected from Canpoint. To make our result more convincing, it is necessary to use problems from other websites for test purpose. Considering the number of problems in PhysQA, we sample 175 problems from 5 types as our test set. Then we use zero-shot and few-shot(prompted by one similar problem) methods to solve them and recalculate the accuracy. All prompted problems come from PhysQA and are calculated by text similarity.

\subsection{Auxiliary Tasks}
To further analyse the model's ability on comprehending the physical scenarios and the knowledge embedded in the problems, we establish several auxiliary tasks on GPT3.5.

\noindent\textbf{Generateing New Physics Word Problems on the Basis of Original Ones} \  We input some of the problems in PhysQA and then instruct the model to generate similar problems.

\noindent\textbf{Summarizing the Core Knowledge in Problems} \  We also selected some problems and tried to ask the model to pick out and summarize the essential knowledge point examined by the problem. If the model could fulfill this task accurately, there is a way of self-prompting that the generated answer could be used as the prompt when solving corresponding problems.  

Since evaluating the equality of the model’s behavior in these tasks quantitatively is challenging, we do not carry them out on all the problems in the dataset. Instead, we randomly sample some problems from each type and generally evaluate their behavior based on sampled the results.

\section{Result}
\subsection{Accuracy of Solving Problems in PhysQA}

When judging whether the model provides an accurate solution, if the value of a relevant physical constant does not appear in the questions (e.g., gravitational acceleration \( g \)), it is acceptable as long as the value used in the calculation approximates the ground truth. For example, it is acceptable to consider \( g \) as either 10N/kg or 9.8N/kg. Regarding the final numeric result of each sub-question, we allow for an error margin of up to 5\%.

The accuracy of different methods is shown in Table \ref{tab2}. We can find that when the prompt of similar problems exists, the accuracy is much more better than zero-shot result. This result show that in solving physics word problems, like many other tasks, the model preforms much better when constructive prompts are available. Besides, There is almost no improvement of the total accuracy when using the textbook as a prompt.It illustrates that the paragraphs of textbook are not strong enough to help model perform better.

\begin{table*}
\centering
\resizebox{\textwidth}{!}
{
\begin{tabular}{ccccccc}
\hline
\textbf{methods} & \textbf{Kinematics(\%)} & \textbf{Mass\&Density(\%)}& \textbf{Mechanics(\%)} & \textbf{Heat(\%)}&\textbf{Electricity(\%)} & \textbf{Total(\%)}\\
\hline
zero-shot & 50.7 &44.8 &45.9 &50.3 &61.7 &49.3 \\
promoted by textbook &46.0 & 40.6 &46.8 &58.0 &60.0 &49.3 \\
prompted by 1 problem &75.9 &74.5 &64.0 &72.9 &85.8 &73.2 \\
prompted by 3 problems &70.0 &75.5 &69.4 &73.2 &71.7 &71.9 \\
\hline
\end{tabular}
}
\caption{
Accuracy of GPT3.5 solving problems in PhysQA. We count the result of every sub-question independently.
}
\label{tab2}
\end{table*}

\subsection{
Reasons for Inaccuracy in Providing a Solution
}

\begin{figure*}[t]
	\subfigure[Calculation error]{
	\includegraphics[width=0.45\textwidth]{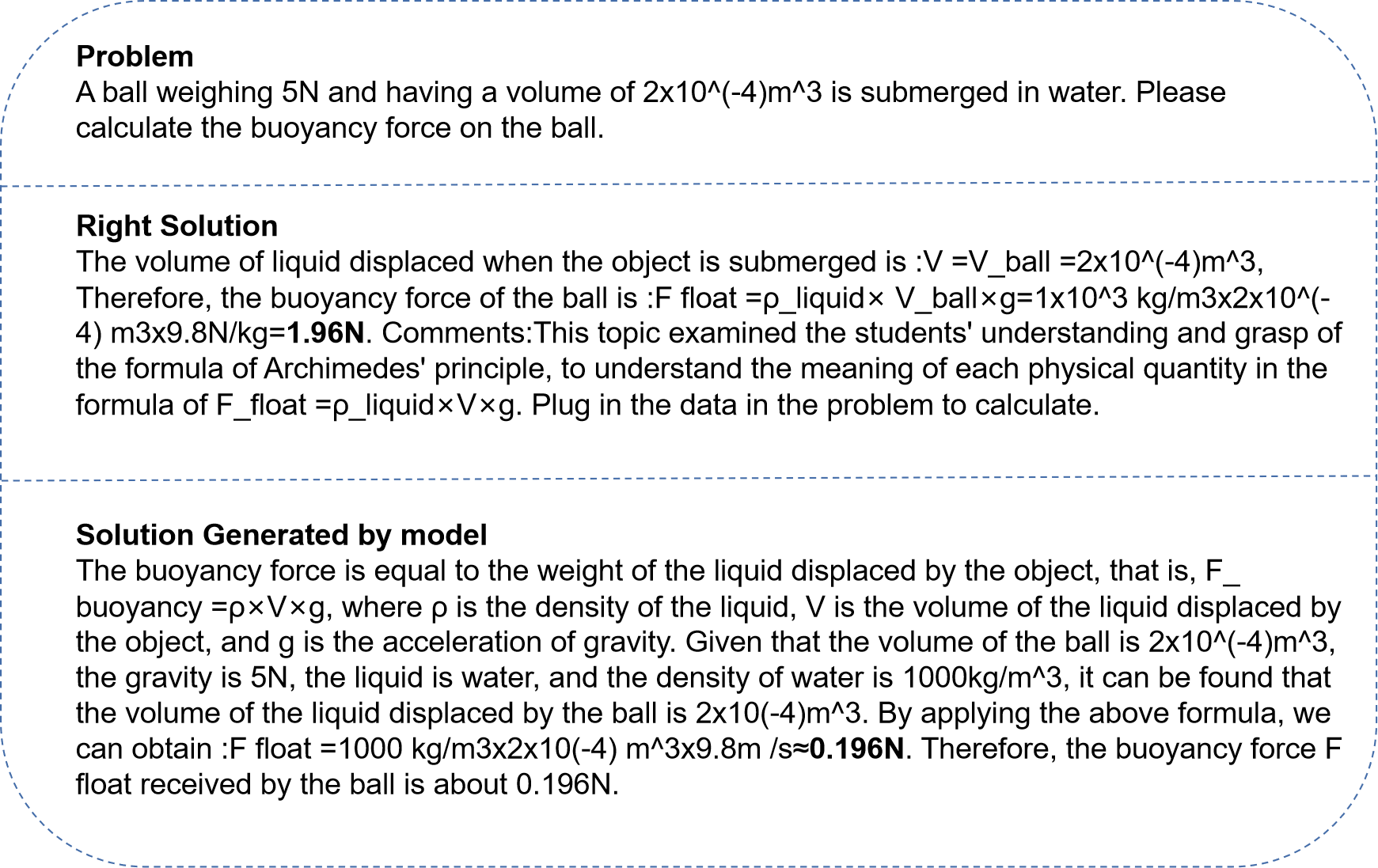} 
        \label{Fig.4(a)}
        }
	\hspace{2mm}
	\subfigure[Unable to understand physical scene]{
	\includegraphics[width=0.45\textwidth]{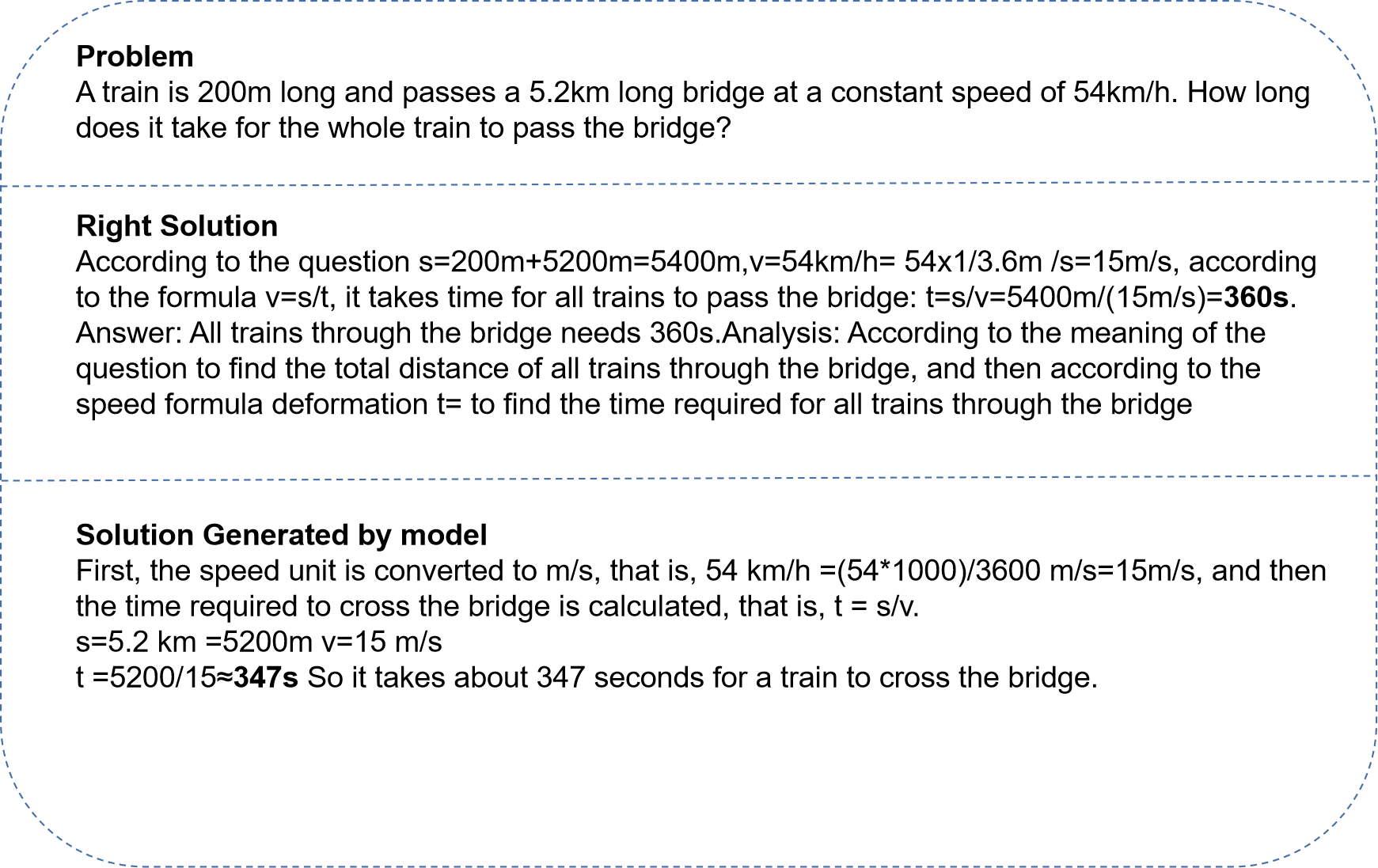} 
        \label{Fig.4(b)}
        }	

	\subfigure[Dimension error]{
		\includegraphics[width=0.45\textwidth]{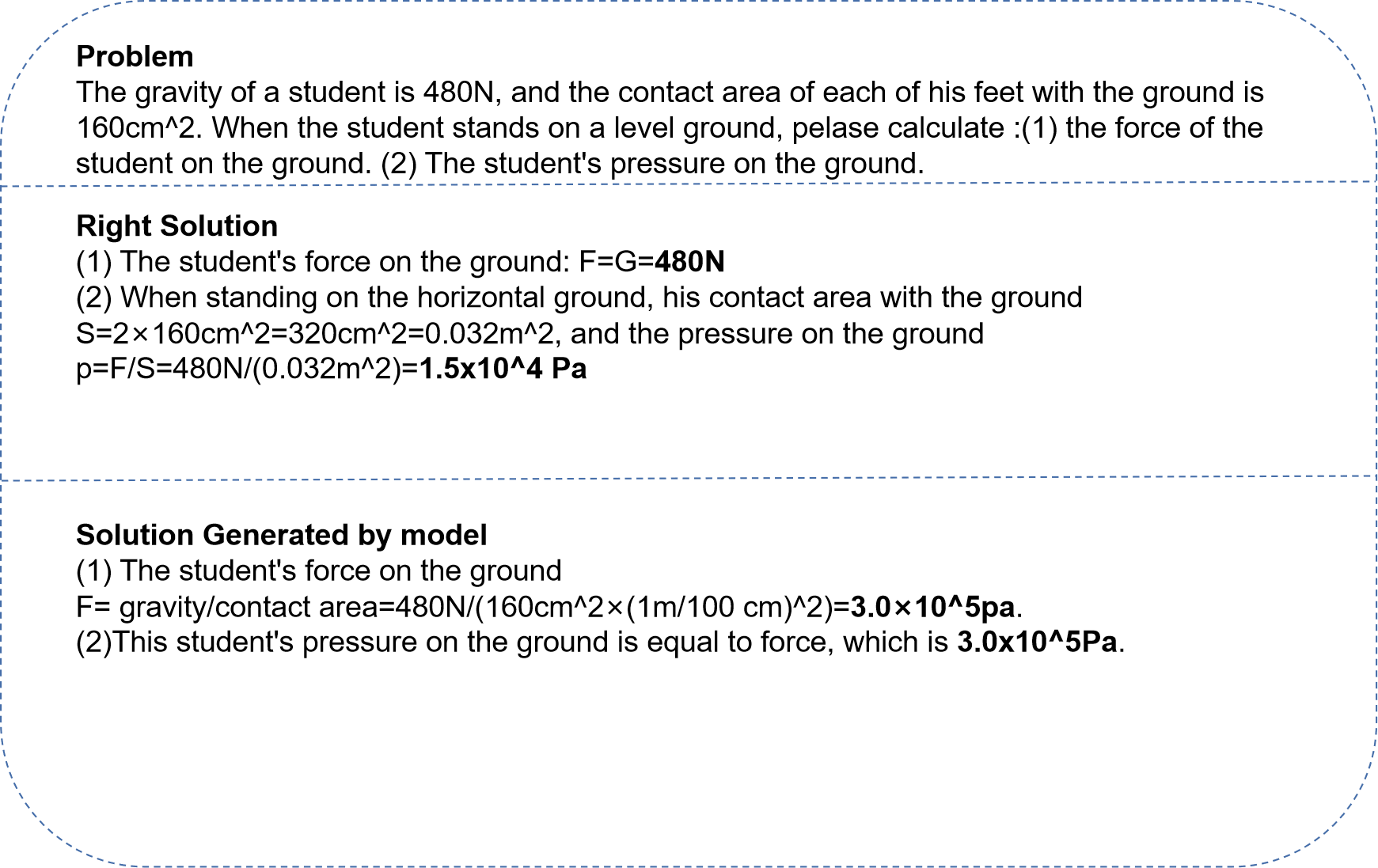} 
        \label{Fig.4(c)}
	}
	\hspace{2mm}
        \subfigure[Unable to understand physical theory and knowledge]{
		\includegraphics[width=0.45\textwidth]{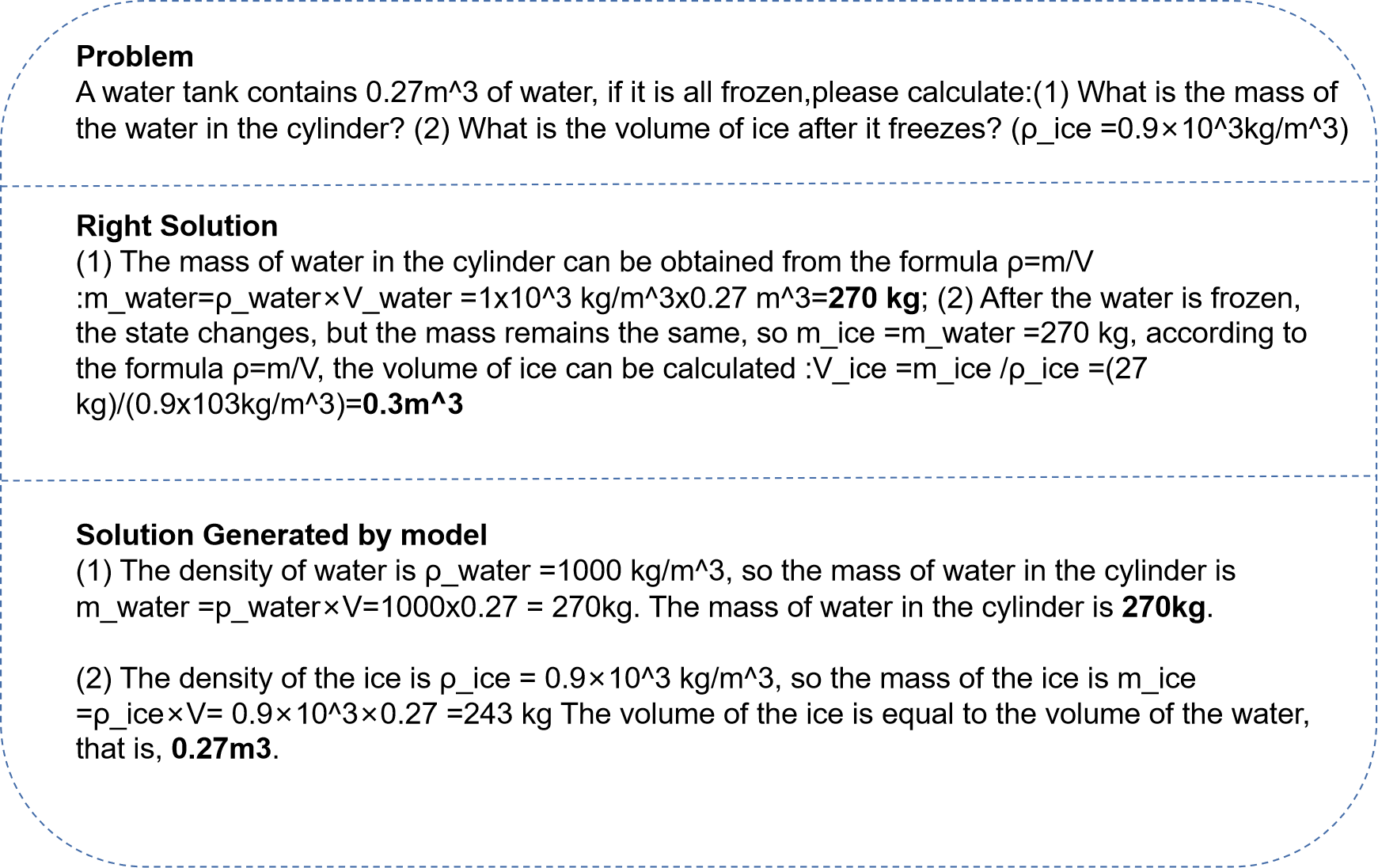} 
        \label{Fig.4(d)}
	}
 
	\caption{Examples of different types of error that appear frequently when solving problems in PhysQA.}
	\end {figure*}

When summarizing results, we found that when model is solving physics word problems, the most frequent types of errors are as follows.
\begin{itemize}
\item \textbf{Calculation Error} \  In some cases, the model would output the perfect correct process of solving the problem except for the numeric result of equation, as seen in Figure \ref{Fig.4(a)} , even when 5\% of maximum error is allowed. This type of error occasionally appears in GPT3.5 and ChatGPT.

\item \textbf{Unable to Understand Physical Scene} \  It means that the model could not fully understand the scene described in problems, resulting in the generation of inaccurate solutions. An example of this error is shown in Figure \ref{Fig.4(b)}. In this problem, the total distance should be equal to the sum of the length of the bridge and the train. However,the model mistakenly regards the length of bridge as distance, which leads to an inaccurate solution.

\item \textbf{Dimension Error} \  Sometimes the model could not understand the dimension of the goal variable. As shown in Figure \ref{Fig.4(c)}, it 
fails to distinguish between pressure and force.

\item \textbf{Unable to Understand Physical Theory and Knowledge} \  As shown in Figure \ref{Fig.4(d)}, during the freezing process, it is actually the mass that remains unchanged. However, the model mistakenly believes that it is the volume that keeps unchanged, which prevent it from solving the problem accurately.
\end{itemize}

\subsection{The Improvement of Accuracy When Prompted by Similar Problems}
\begin{figure}
	\subfigure[]{
	\includegraphics[width=0.45\textwidth]{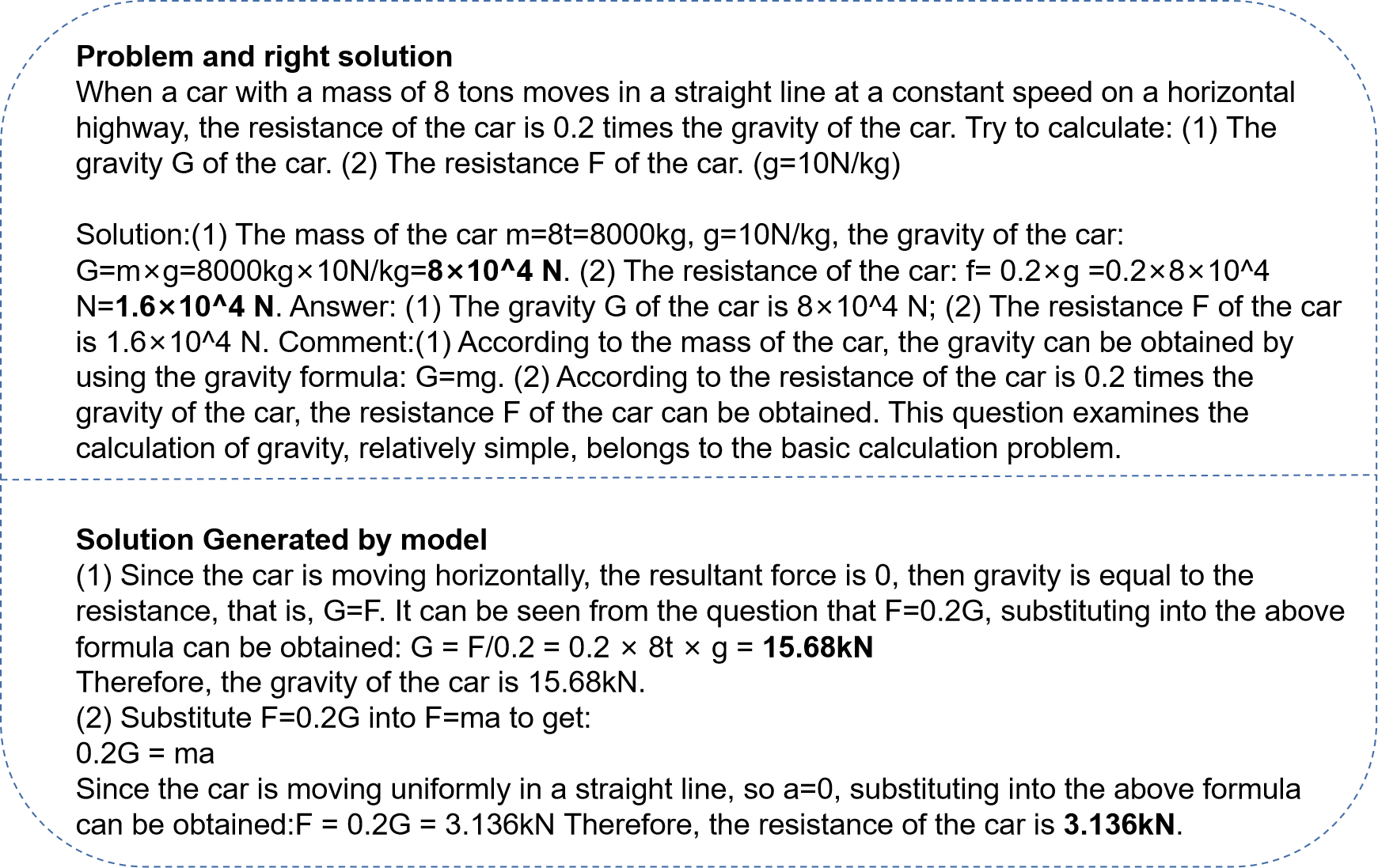} 
        \label{Fig.5(a)}
        }

	\subfigure[]{
		\includegraphics[width=0.45\textwidth]{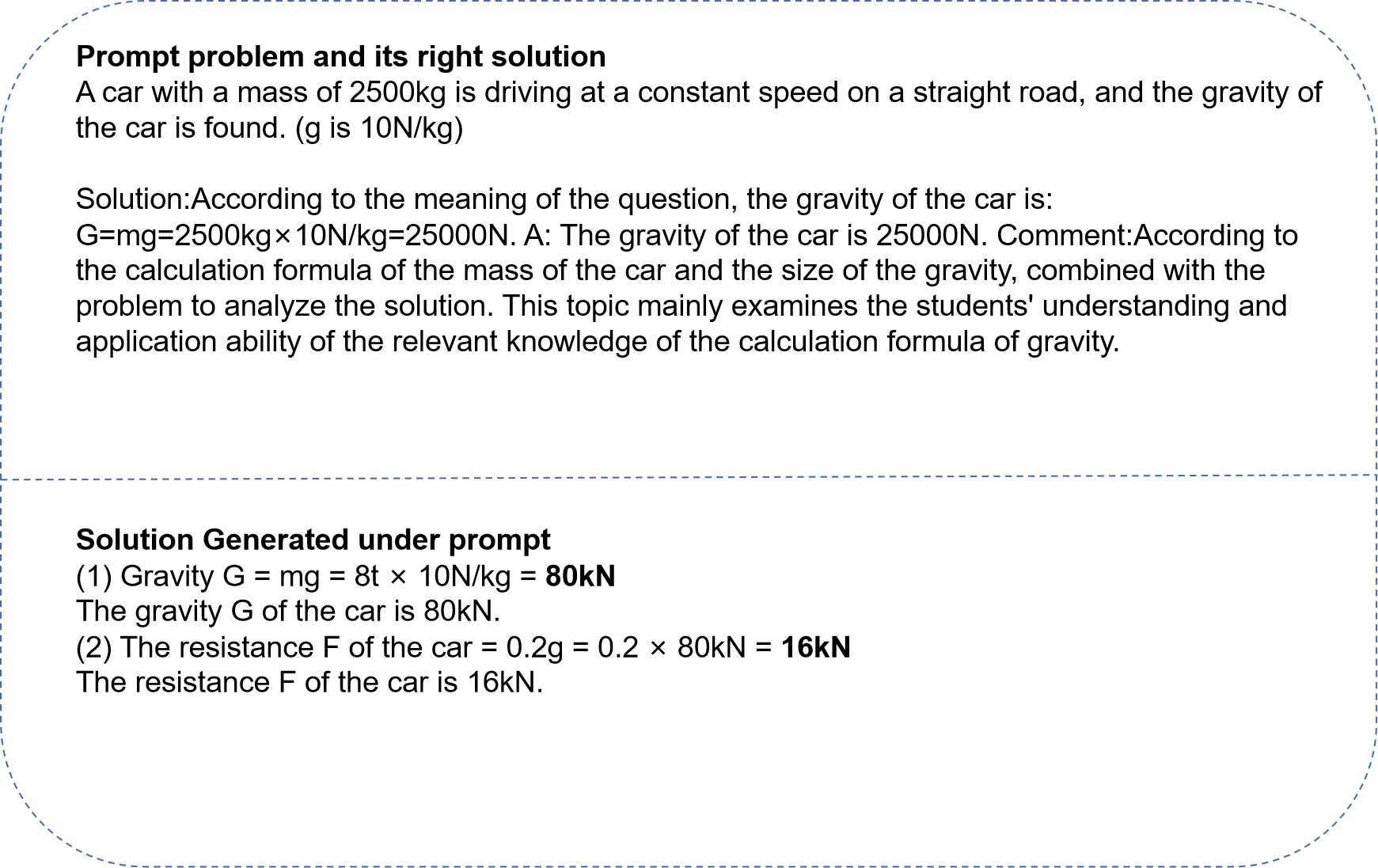} 
        \label{Fig.5(b)}
	}

	\caption{Example of using similar problem as prompt to help model solve the problem accurately.}
        \label{fig5}
	\end {figure}

When provided with prompts containing similar problems and their corresponding answers, the model's performance is significantly better than in other situations. The determination of the 'most similar question' in the dataset is based on text similarity, which means that similar problems typically involve analogous scenarios and the same physical knowledge points. Some of the solutions obtained from websites also include analyses and comments related to the problems, and these can also enhance the quality of the prompts. By consulting relevant problems and solutions, as demonstrated in Fig \ref{fig5}, the model gains a better grasp of the physical context, improving its ability to apply relevant knowledge accurately and produce the correct answer.

We also notice that when inputting all three similar problems as prompts, the total accuracy is even slightly lower than inputting just one problem as prompt. A possible explanation is that the text of the three problems and their solutions is quite lengthy, which sometimes causes the model to mix up the conditions and sub-questions between the prompts and the problems to be solved. This result also shows that when solving physics word problems, more prompts do not necessarily lead to better performance.

The results of solving problems in our test set also demonstrate the positive impact of prompt questions. Given the small size of the test set, we are not focused on the absolute accuracy on the test set but rather on the improvement in accuracy relative to zero-shot learning. As indicated in Table \ref{tab3}, the presence of prompt problems leads to an improvement in accuracy of approximately 10\%. 

\begin{table}
\centering
{
\begin{tabular}{cc}
\hline
\textbf{methods} & \textbf{Total(\%)}\\
\hline
zero-shot&52.6 \\
prompted by 1 problem &63.0 \\
\hline
\end{tabular}
}
\caption{
Accuracy of GPT3.5 solving problems in test set. We count the result of every sub-question independently.
}
\label{tab3}
\end{table}
\subsection{The Ability of GPT3.5 in Explaining, Generating and Summarizing Physics Word Problems}

Compared with traditional MWP solver, an advantage of LLM is that its  interpretability is much better since its output not only contains the equation and numeric result, but also includes the essential explanation and train of thought, which is helpful in assisting students and users to deeply understand this problem.

\begin{figure}[t]
\centering
\includegraphics[width=0.45\textwidth]{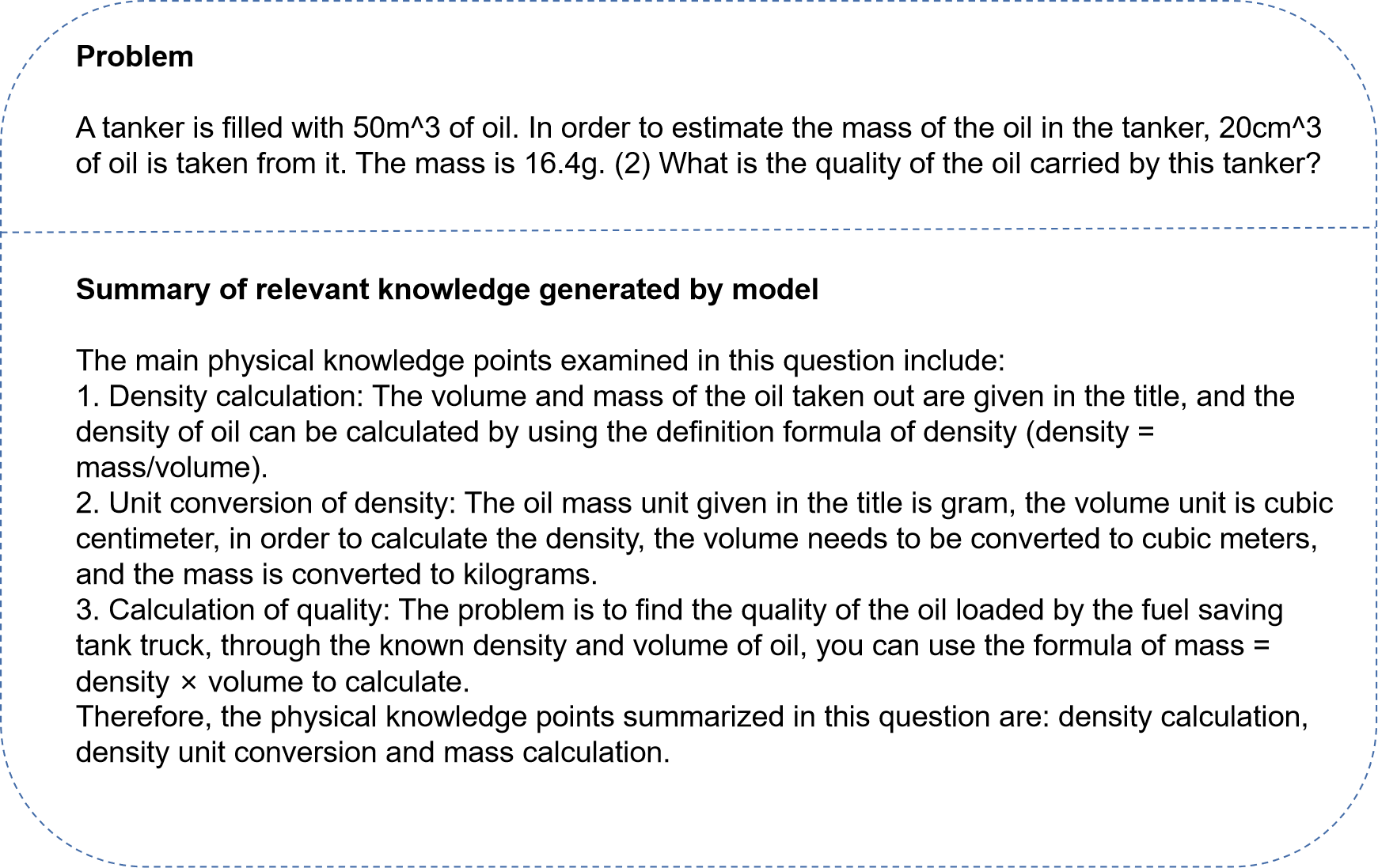}
\caption{An example of asking model to generate the summary of problem.}
\label{fig6}
\end{figure}

Additionally, during the auxiliary tasks, we have also observed that when we input questions to the model and ask it to generate similar questions based on them, in most cases, it is able to generate valid new problems. This capability could simplify the task of teachers in creating fresh questions or papers for their students. And 
instructing the model to identify core knowledge points, we find that though it list up most of essential knowledge or key formula contained, some of the details in it are not accurate enough, which might explain the reason why generated solution goes wrong in zero-shot learning.(An example of summary is shown in Figure \ref{fig6}) It shows that although GPT3.5 may not always solve the problem accurately, at least it can be used as a tools to assist the user in obtaining relevant knowledge point, which can aid humans in arriving at the correct solution.

\subsection{Limitations}
Our work mainly includes the following limitations:

\noindent\textbf{The uncertainty of LLM} \  If we ask GPT3.5 a same question twice or more, its output result might differ. Moreover, we found that when using it to solve problems, the model's performance fluctuates. Regardless of the difficulty of the problems, there are times when the model completely fails to solve the problem for over seven consecutive questions, and then its accuracy returns to a normal level. This phenomenon occurs less frequently when using similar problems as prompt.

\noindent\textbf{The scale of dataset} \  Compared to large-scale math word problem datasets containing over 20K problems, our dataset is much smaller in scale. The advantage of a larger dataset lies in its ability to yield more persuasive results. More importantly, a larger dataset offers a higher likelihood for target problems to find similar problems within it as prompts, potentially enhancing the accuracy of few-shot learning methods.

\section{Conclusion}
We firstly collected and annotated PhysQA, which is the first dataset of junior high school physics word problems at a scale of 1K. This dataset can also be valuable for future research on the automatic solution of physics word problems. Our work demonstrates that GPT3.5 has the capability in solving and explaining junior-high-school-level physics word problems in the PhysQA dataset, as well as generating new problems based on them. The accuracy achieved when using similar problems as prompts is significantly higher than the result of zero-shot learning and prompts based on a textbook. This highlights the importance of effective prompts in solving physics word problems.

\section{Ethical Impacts}
The problems in PhysQA are collected from educational and test paper websites and they are only used for academic research. The copyright of problems belong to original websites.

\section*{Acknowledgements}
This work is supported by Department of Physics, Fudan University and Physics Teaching Lab, Fudan University.

\bibliography{anthology,custom}
\bibliographystyle{acl_natbib}

\appendix

\end{document}